\documentclass[letterpaper, 10 pt, journal, twoside]{IEEEtran}

\usepackage{graphicx}
\usepackage{tikz}
\usepackage{comment}
\usepackage{amsmath,amssymb} % define this before the line numbering.
\usepackage{color}
\usepackage{colortbl}
\usepackage{multicol}
\usepackage{multirow}
\usepackage{cite}
\usepackage{booktabs}

\usepackage[accsupp]{axessibility}  % Improves PDF readability for those with disabilities.

\usepackage{url}
\usepackage{hyperref}

\newcommand{\etal}{\textit{et al}. }
\newcommand{\ie}{\textit{i}.\textit{e}., }
\newcommand{\eg}{\textit{e}.\textit{g}. }

\def\ralhigh#1{{#1}}  % <- un-comment this line for a non-highlighted version

\title{
RFTrans: Leveraging Refractive Flow of Transparent Objects for Surface Normal Estimation and Manipulation
}

\author{Tutian Tang$^{1*}$, Jiyu Liu$^{1*}$, Jieyi Zhang$^{1}$, Haoyuan Fu$^{1}$, Wenqiang Xu$^{1}$ and Cewu Lu$^{2}$% <-this % stops a space
\thanks{Manuscript received: October 20, 2023; Revised December 12, 2023; Accepted January 29, 2024.}%Use only for final RAL version
\thanks{This paper was recommended for publication by Editor Markus Vincze upon evaluation of the Associate Editor and Reviewers' comments.
This work was supported by the National Key R\&D Program of China (No. 2021ZD0110704), Shanghai Municipal Science and Technology Major Project (2021SHZDZX0102), Shanghai Qi Zhi Institute, and Shanghai Science and Technology Commission (21511101200)} %Use only for final RAL version
\thanks{*Equal contribution.} %
\thanks{$^{1}$Tutian Tang, Jiyu Liu, Jieyi Zhang, Haoyuan Fu and Wenqiang Xu are with School of Electronic Information and Electrical Engineering, Shanghai Jiao Tong University, Shanghai, China
    {\tt\footnotesize \{tttang, waterloo-sunset, yi\_eagle, simon-fuhaoyuan, vinjohn\}@sjtu.edu.cn}} %
\thanks{$^{2}$Cewu Lu is the corresponding author, a member of Qing Yuan Research Institute and MoE Key Lab of Artificial Intelligence, AI Institute, Shanghai Jiao Tong University, Shanghai, China
    {\tt\footnotesize lucewu@sjtu.edu.cn}}%} %
\thanks{Digital Object Identifier (DOI): see top of this page.}
}

\begin{document}

\maketitle
% \thispagestyle{empty}
% \pagestyle{empty}

%%%%%%%%%%%%%%%%%%%%%%%%%%%%%%%%%%%%%%%%%%%%%%%%%%%%%%%%%%%%%%%%%%%%%%%%%%%%%%%%
\begin{abstract}

Transparent objects are widely used in our daily lives, making it important to teach robots to interact with them. However, it's not easy because the reflective and refractive effects can make depth cameras fail to give accurate geometry measurements. To solve this problem, this paper introduces RFTrans, an RGB-D-based method for surface normal estimation and manipulation of transparent objects.
By leveraging refractive flow as an intermediate representation, the proposed method circumvents the drawbacks of directly predicting the geometry (\eg surface normal) from images and helps bridge the sim-to-real gap.
It integrates the RFNet, which predicts refractive flow, object mask, and boundaries, followed by the F2Net, which estimates surface normal from the refractive flow. To make manipulation possible, a global optimization module will take in the predictions, refine the raw depth, and construct the point cloud with normal. An off-the-shelf analytical grasp planning algorithm is followed to generate the grasp poses.
We build a synthetic dataset with physically plausible ray-tracing rendering techniques to train the networks. Results show that the proposed method trained on the synthetic dataset can consistently outperform the baseline method in both synthetic and real-world benchmarks by a large margin.
Finally, a real-world robot grasping task witnesses an 83\% success rate, proving that refractive flow can help enable direct sim-to-real transfer.
The code, data, and supplementary materials are available at \url{https://rftrans.robotflow.ai}.

\end{abstract}


%%%%%%%%%%%%%%%%%%%%%%%%%%%%%%%%%%%%%%%%%%%%%%%%%%%%%%%%%%%%%%%%%%%%%%%%%%%%%%%%
\section{Introduction}
\IEEEPARstart{T}{ransparent}
objects like glass goblets and plastic bottles are widely used in our daily lives. While humans can easily interact with transparent objects, teaching robots to manipulate them is not straightforward. One of the main barriers lies in the perception part, for transparency implies several special physical properties, including reflection, refraction, and the absence of color and texture. These properties often cause most 3D sensors, including LiDAR and RGB-D cameras, to fail to produce accurate geometry measurements for transparent objects~\cite{trans_survey}. Consequently, mainstream methods~\cite{a4t,cleargrasp} for transparent object manipulation typically adopt a two-stage pipeline. In the first stage, neural networks are used to recover the geometry of transparent objects based on noisy and inaccurate readings from commercial RGB-D cameras. Then, in the second stage, manipulation algorithms can be applied to the point cloud derived from the estimated geometry. Since these two stages are carried out sequentially, ensuring high-quality reconstructed geometry becomes crucial for reliable manipulation.

\begin{figure}
    \centering
    \includegraphics[width=\columnwidth]{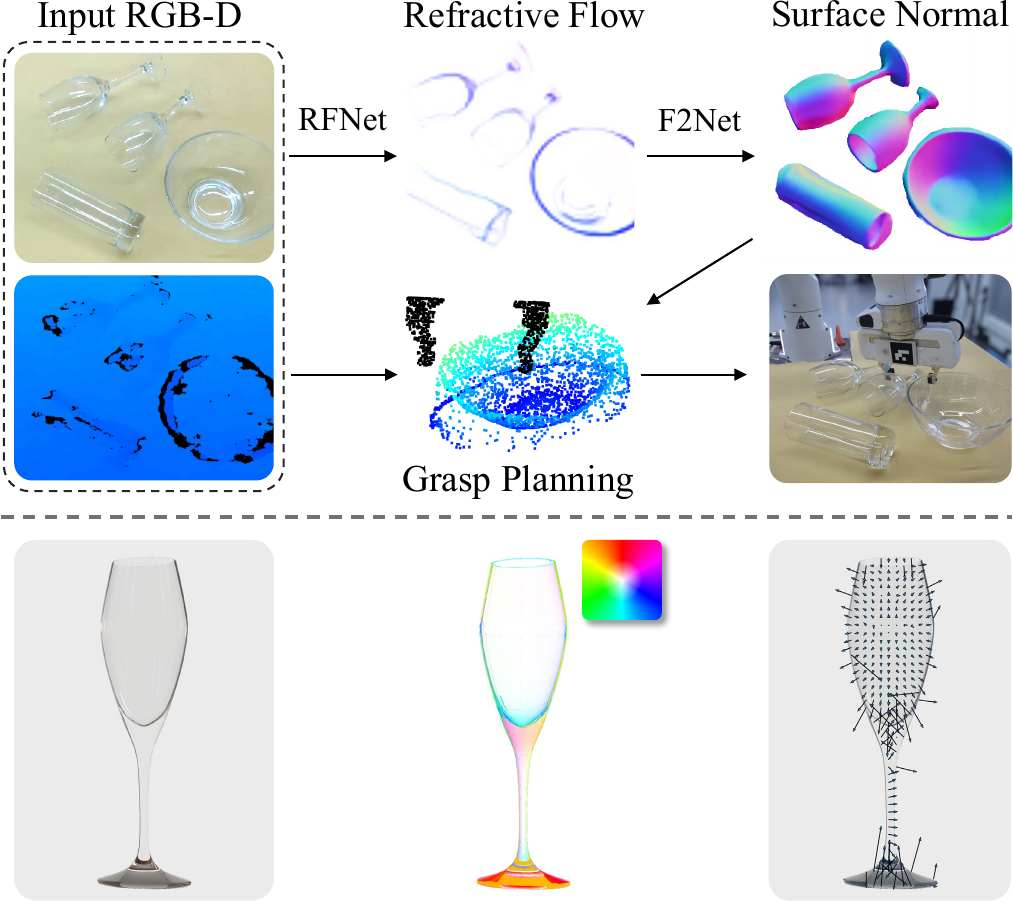}
    \caption{
        \textbf{Top:} Transparency can cause inaccurate and missing depth captured by those widely-used RGB-D cameras. We utilize refractive flow to recover the surface normal and finally get the point cloud for robot manipulation.
        \textbf{Bottom:}
            \textbf{(Left)} A common wine glass.
            \textbf{(Middle)} We visualize the refractive flow by color map. The color represents the direction and magnitude. White indicates no refraction on the pixel.
            \textbf{(Right)} We sample some points on the image and show the corresponding refractive flow as arrows, which start from the foreground pixels on the glass to their corresponding pixels on the background.
    }
    \label{fig:teaser}
\end{figure}

The recovery of transparent object geometry has been studied for decades. Previous approaches mostly relied on tracking the delicate light path using images from multiple viewpoints~\cite{PositionNormalConsistency,FixedViewPoint,Full3DRecon}. ClearGrasp~\cite{cleargrasp} pioneered single image-based surface normal estimation, which directly predicts the surface normal from the RGB images. However, such direct prediction can be problematic in two aspects.
First, the correlation between RGB patterns and the underlying geometry is insignificant, especially when the background is complex.
Second, obtaining the accurate surface normal in the real world is hard, so researchers must generate large amounts of synthetic data to train and evaluate the networks. In this case, the sim-to-real transferability becomes a challenging problem.

In this work, we propose \textbf{RFTrans}, which achieves surface normal estimation and manipulation of transparent objects based on a single RGB-D image. The main character distinguishing RFTrans from other works is that it uses the physical property, refraction, by adopting refractive flow~\cite{tom-net} as an intermediate representation to mitigate the challenges of predicting surface normal directly from RGB images.
Illustrated in Figure~\ref{fig:teaser}, the refractive flow is a per-pixel offset map between the transparent foreground and non-transparent background pixels to model the refractive effect of transparent objects. On each pixel, the refractive flow forms a 2D offset vector. We find refractive flow a good intermediate representation, for it features several merits, such as \textit{small sim-to-real gap}, \textit{stable under different ambient lights}, and \textit{insensitive to complex background}. The details will be discussed in Section~\ref{sec:rf_property}.

In RFTrans, the \textbf{RFNet} first predicts refractive flow, object mask, and boundary based on RGB images. The \textbf{F2Net} then estimates the surface normal based on refractive flow. Then, these geometry-related elements, along with the original depth map, will be fed into a global optimization module as in~\cite{cleargrasp}, which will output the point cloud of transparent objects with normal to be fed into an off-the-shelf grasp planning algorithm. Here, we use ISF~\cite{isf}, an analytical grasp planning algorithm built on top of a heuristic surface matching metric. It can generate the grasp pose and guide the robot's execution.

To train and evaluate the proposed method, we build a synthetic dataset of 62 transparent objects with RFUniverse~\cite{rfu}, which features the latest simulation and rendering technologies and can thus generate physically plausible images. In addition, we directly test our model trained with synthetic data in a real-world benchmark~\cite{cleargrasp}. Results show that our method outperforms the baseline method consistently. In the real-world grasp task, the success rate increases to 83\% from 35\% after the proposed method is applied.

Our main contributions can be summarized as follows:
\begin{itemize}
    \item We propose RFTrans, a pipeline for surface normal estimation and manipulation of transparent objects based on RGB-D images. Refractive flow is used as an intermediate representation to help reconstruct accurate surface geometry.

    \item We construct a synthetic dataset of 62 transparent objects. The data generation pipeline is fully open-sourced.

    \item We set up a real-world grasping task to prove that RFTrans can enable direct sim-to-real transfer.
\end{itemize}

\begin{figure*}
    \centering
    \includegraphics[width=0.9\linewidth]{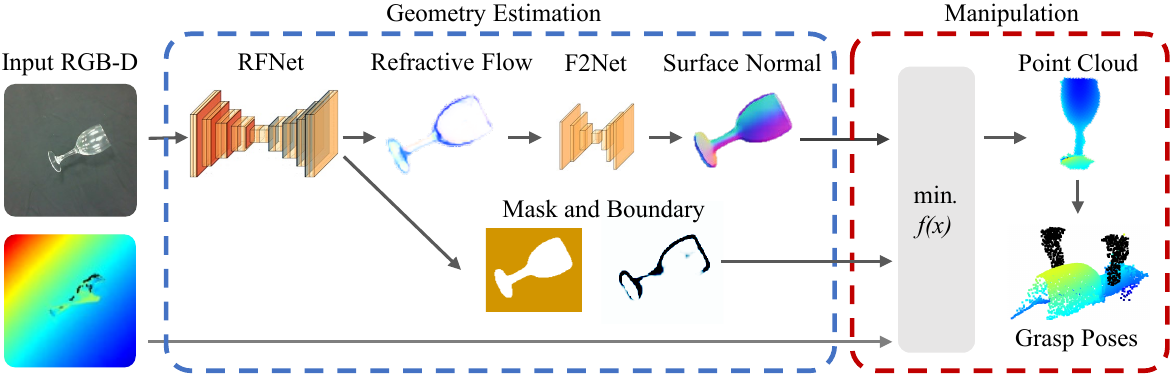}
    \caption{Given an RGB-D image, RFNet first predicts the mask, the boundary, and the refractive flow of transparent objects. Next, F2Net will predict the surface normal based on the refractive flow. The global optimization will generate the singulated point cloud with normal. Finally, we apply the off-the-shelf manipulation algorithm, ISF, to generate grasp poses. The black points represent the fingers of the Franka Emika Panda robot.}
    \label{fig:pipeline}
\end{figure*}

%%%%%%%%%%%%%%%%%%%%%%%%%%%%%%%%%%%%%%%%%%%%%%%%%%%%%%%%%%%%%%%%%%%%%%%%%%%%%%%%

\section{Related Work}

Our proposed method is most closely related to those approaches that utilize the refractive properties for transparent object geometry estimation and manipulation.

\subsection{Refractive Property Estimation for Transparent Objects}

Refractive flow describes the refractive properties of transparent objects. Initial work in this domain~\cite{139539, agarwal_refractive_2004, 1544905, 6126533} involved the reconstruction of water surfaces by positioning a pre-defined pattern beneath a water tank and leveraging an optical flow-based algorithm to find the corresponding points between the camera pixels and the points from the pattern.
Later methods~\cite{6126367,1544889} remove the dependence on the water tank but impose some assumptions on the geometry of the objects.
Some other correspondence-based methods~\cite{xu_hybrid_2023,gas_flow_light_path_approx} measure the refractive flow without any prior of the geometry of the transparent objects.
A more recent work, TOM-Net~\cite{tom-net}, estimates the refractive flow with neural networks trained on synthetic data, but it's for image composition and matting instead of geometry recovery.
There are many different methods to get refractive flow\cite{1544889,6126367,xu_hybrid_2023}.
In this work, we model the refractive flow in a gray code-based approach~\cite{lyu_differentiable_siggraph} for its ease of use.

\subsection{Estimating Geometry of Transparent Objects}

Estimating the surface geometry of transparent objects has long been a challenge~\cite{trans_reconstruction_2010_ihrke,trans_survey}.
Some methods~\cite{10.1145/1179849.1179918,10.1145/1360612.1360686,10.1145/3072959.3073693} necessitate direct, intrusive interaction with the objects, keeping them unsuitable for robotic manipulation tasks. Murali \etal~\cite{murali_touch_2023} introduce the non-intrusive tactile modality for reconstruction without damaging the objects.

Conversely, vision-based methods exploit the phenomena of reflection and refraction to deduce the underlying geometric attributes.
Since specular reflection usually only happens in a small area of the transparent object body, reflection-based approaches traditionally either involve much human labor~\cite{4408882}, require the periodic movement of the lighting source~\cite{5995472}, or use multiple camera views~\cite{1544905}.
Refraction, however, can be observed through nearly the whole body of a transparent object, providing much more information about surface geometry than reflection. Therefore, Kutulakos and Steger~\cite{1544889} utilize a combination of both phenomena to help reconstruct the objects.

The past decade has seen tremendous progress in neural networks. Stets \etal~\cite{8659154} and Sajjan \etal~\cite{cleargrasp} are the pioneers of applying deep learning techniques in this field. Both methods utilize neural networks to estimate the surface normal of transparent objects directly from a single RGB image.
Later, researchers push these methods into end-to-end depth restoration pipelines by using CNN-based networks~\cite{fang2022transcg}, transformer-based networks~\cite{dai_domain_2022}, or implicit representations~\cite{implicit_depth}.
In multi-view vision systems, a recent trend is to adopt neural radiance fields (NeRF) to represent the scenes and objects implicitly. Li \etal~\cite{li_through_2020} are the pioneers in applying NeRF in transparent object reconstruction, followed by Dex-NeRF~\cite{ichnowski_dex-nerf_2021}. However, a minimum of $3\times3$ camera array system is required, which hinders its application in robotics. Kerr \etal~\cite{kerr_evo-nerf_2022} further improve its speed and remove the need for the camera array. Dai \etal~\cite{Dai2023GraspNeRF} propose GraspNeRF, which leverages generalizable NeRFs to reduce the number of images required to reconstruct one single scene. However, the NeRF-based methods still require multiple sparse-view RGB images to reconstruct the objects. All these methods consider the relationship between RGB images and the surface geometry as a black box without explicitly exploiting the phenomena of reflection and refraction. Our method first adopts the refractive flow as an intermediate representation, making it less sensitive to backgrounds (Sec.~\ref{sec:rf_property_complex_back}), less data-hungry (Sec.~\ref{sec:experiment_depth}), and have better sim-to-real transferability (Sec.~\ref{sec:experiment_tab_1}).

\subsection{Transparent Object Manipulation}

Transparent object manipulation is an important application area for geometry estimation. The upstream geometry estimation and the downstream manipulation algorithm can be either tightly or loosely coupled. Apparently, a loose coupling design allows the whole manipulation framework to benefit from the latest advances from both sides easily. ClearGrasp~\cite{cleargrasp} is a typical loose coupling framework, followed by Jiang \etal in A4T~\cite{a4t}, which sets a good example of bridging depth completion and manipulation via affordance. Later, Fang \etal~\cite{fang2022transcg} and Dai \etal~\cite{dai_domain_2022} witness the direct improvement by applying the latest CNNs and transformers in depth completion. In Dex-NeRF~\cite{ichnowski_dex-nerf_2021}, although the scene is represented implicitly via NeRF, the depth map is still extracted explicitly, leading the whole framework into a loose coupling manner. However, GraspNeRF~\cite{Dai2023GraspNeRF} shows the tight coupling way, which introduces the Truncated Signed Distance Field (TSDF) as the bridge between implicit scene representation and grasping. There are also many data-driven approaches~\cite{Weng-2020-123091,cao2021fuzzy,9636459} aimed at directly generating grasping poses for transparent objects from noisy RGB-D images without explicit geometry estimation or depth completion. These deep-learning-based manipulation algorithms are usually data-hungry and not flexible enough to be a downstream grasp planner. First, they may fail to generalize between different models of RGB-D cameras. Second, if the user wants to add some new objects or adopt a new gripper of a different configuration, the networks must be retrained. Therefore, we decide to go for an analytical grasp planning algorithm, ISF~\cite{isf}, for the real-world grasping task.

%%%%%%%%%%%%%%%%%%%%%%%%%%%%%%%%%%%%%%%%%%%%%%%%%%%%%%%%%%%%%%%%%%%%%%%%%%%%%%%%

\section{Method}
\label{sec:method}

In this section, we will first give the definition and acquisition method of refractive flow (Sec.~\ref{sec:rf_def_and_acq}). Next, we discuss some properties of the refractive flow to show the reason why we adopt it as the intermediate representation (Sec.~\ref{sec:rf_property}). Then, we describe how to estimate the geometry of transparent objects with neural networks (Sec.~\ref{sec:method_pipeline}). To train and evaluate the neural networks, we construct a synthetic dataset based on RFUniverse~\cite{rfu} (Sec.~\ref{sec:method_dataset}). Finally, we will introduce our real-world grasping system in Section~\ref{sec:method_grasp}. The whole pipeline is illustrated in Figure~\ref{fig:pipeline}.

\subsection{Refractive Flow: Definition and Acquisition}
\label{sec:rf_def_and_acq}

\begin{figure}
    \centering
    \includegraphics[width=\columnwidth]{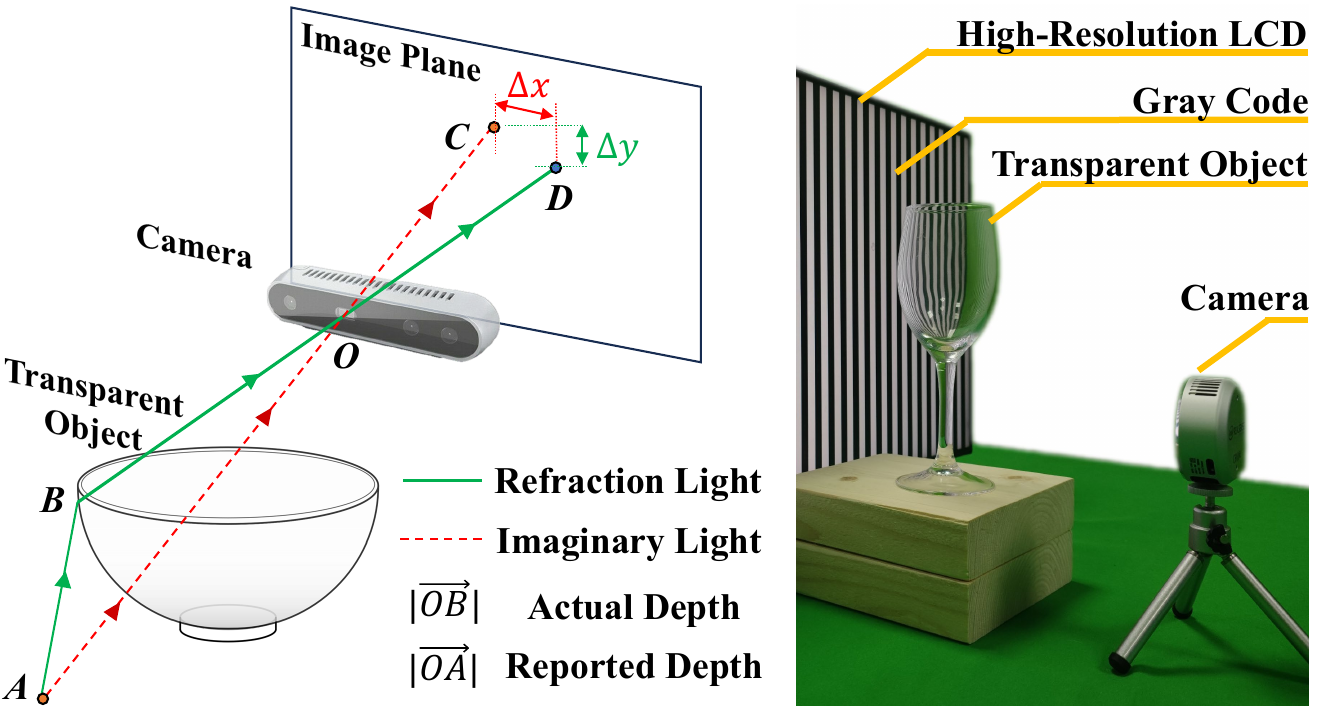}
    \caption{
        \textbf{Left:} Point $O$ is the optical center of the pin-hole camera. Point $A$ is the point on the non-transparent background, \eg a table. The refractive effect takes place at point $B$. $\overrightarrow{AB}$ is the incident ray and $\overrightarrow{BO}$ is the refracted ray. Point $D$ is the image of $A$ on the image plane. Without the transparent object, an imaginary ray will be directly from $A$ to $O$, intersecting the image plane at point $C$. The orthogonal distances between $C$ and $D$ on the image plane, $(\Delta x, \Delta y)$, is the refractive flow at point $D$. Due to transparency, when the Type II error happens, RGB-D cameras usually report $proj|\overrightarrow{OA}|$ as the depth value of point $D$, while the actual depth should be $proj|\overrightarrow{OB}|$, where $proj|\cdot|$ denotes the projected length on the principal axis.
        \textbf{Right:} The data acquisition system to capture refractive flow.
    }
    \label{fig:flow_definition}
\end{figure}

RGB-D cameras usually produce two kinds of errors on transparent objects~\cite{trans_survey}. Type I error happens when reflection happens and the camera fails to detect the depth value, resulting in incomplete depth maps. Type II error is highly related to the refractive effect, which occurs when the light refracts through the object's surface and is reflected back by the non-transparent background. Figure~\ref{fig:flow_definition} illustrates the relationship between refraction and Type II error.

Refractive flow is used to model the refractive effect of a transparent object. Intuitively, each \textit{pixel} on the refractive flow is a 2D vector $(\Delta x, \Delta y)$, which indicates the offset between the foreground pixel and its refraction correspondence on the background image~\cite{tom-net}. For those commonly-used thin-shell transparent objects, we can usually observe significant refraction near the edges. Therefore, the magnitude of the refractive flow is usually large near the boundaries, while the direction depends on a lot of factors including the structure of the object and the direction of the camera.

We use the gray code-based calibration method~\cite{lyu_differentiable_siggraph} to acquire the refractive flow. As illustrated in Figure~\ref{fig:flow_definition}, the transparent object is placed between a high-resolution LCD monitor and a fixed camera. The LCD monitor displays a sequence of 20 gray-coded images, including 10 vertical and 10 horizontal patterns. The camera is calibrated before the acquisition process~\cite{zhang_calibration}, and it captures the corresponding images for the calibration process to get the refractive flow. Strong direct light should be avoided to prevent large reflective areas on the object. Additionally, for generating synthetic data, the system can be cloned into a digital twin in simulation, ensuring a small sim-to-real gap, as discussed in Section~\ref{sec:rf_property_sim2real}.

\subsection{Properties of Refractive Flow}
\label{sec:rf_property}

Apparently, the refractive flow will change according to the transparent objects' viewpoint and geometry. Therefore, it can encode rich information about the geometry.
We find refractive flow features several merits to be a good intermediate representation for neural networks.

\subsubsection{Small Sim-to-Real Gap}
\label{sec:rf_property_sim2real}
\begin{figure}
    \centering
    \includegraphics[width=0.95\columnwidth]{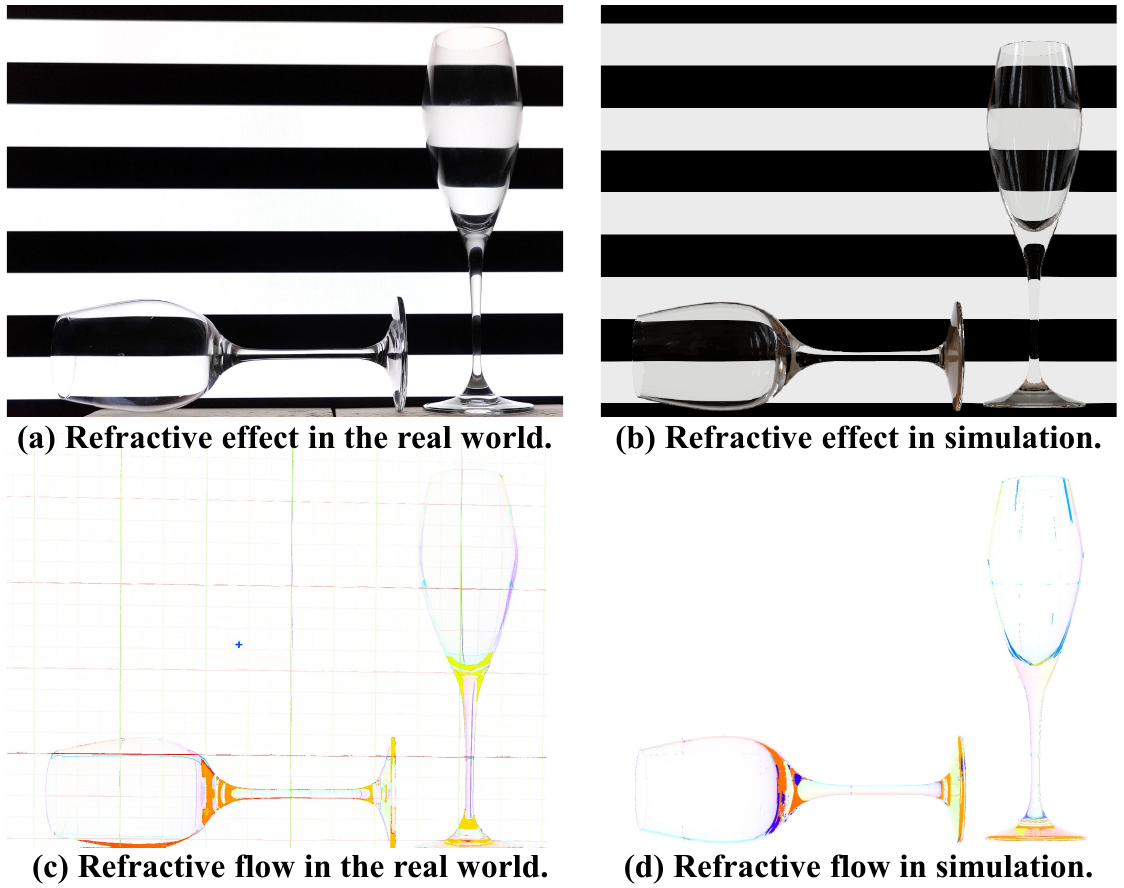}
    \caption{The refractive effect and the corresponding refractive flow in the real world and our simulation environment.}
    \label{fig:sim-to-real-refractive-flow}
\end{figure}

We adopt the gray-code calibration process to generate the refractive flow. Thanks to the latest ray tracing-based rendering technology, we can get photo-realistic and physically plausible images. Figure~\ref{fig:sim-to-real-refractive-flow} shows one of the required images in the calibration process and the resulting refractive flow, both in simulation and in the real world. The sim-to-real gap is small. Also, we will show metrics on a real-world benchmark later in Section~\ref{sec:experiment_tab_1}.

\subsubsection{Stable Under Different Ambient Light}
\label{sec:rf_property_light}
According to Snell's law~\cite{feynman1965flp}: $v_2\sin{\theta_1}=v_1\sin{\theta_2}$, where $v_1, v_2$ are the phase velocities of light in two different media, and $\theta_1, \theta_2$ are the incidence and refraction angles respectively, we can conclude that different ambient lights can cause the gray code pattern shift.
To quantitatively inspect the shift, we put a camera, an LED lamp, and a wine glass in front of a checkerboard. The color temperature of the LED lamp can be tuned from 2600K to 5000K. We compare the rooted mean squared error (\textbf{RMSE}) in pixels of the refractive patterns under the same viewpoint with different color temperatures. Results show that the RMSE value always stays around $0.1$px on a change of color temperature. Therefore, the refractive flow calculated based on these patterns should also be stable. For details, please refer to the supplementary video.

\subsubsection{Insensitive to Complex Background}
\label{sec:rf_property_complex_back}
\begin{figure}
    \centering
    \includegraphics[width=\columnwidth]{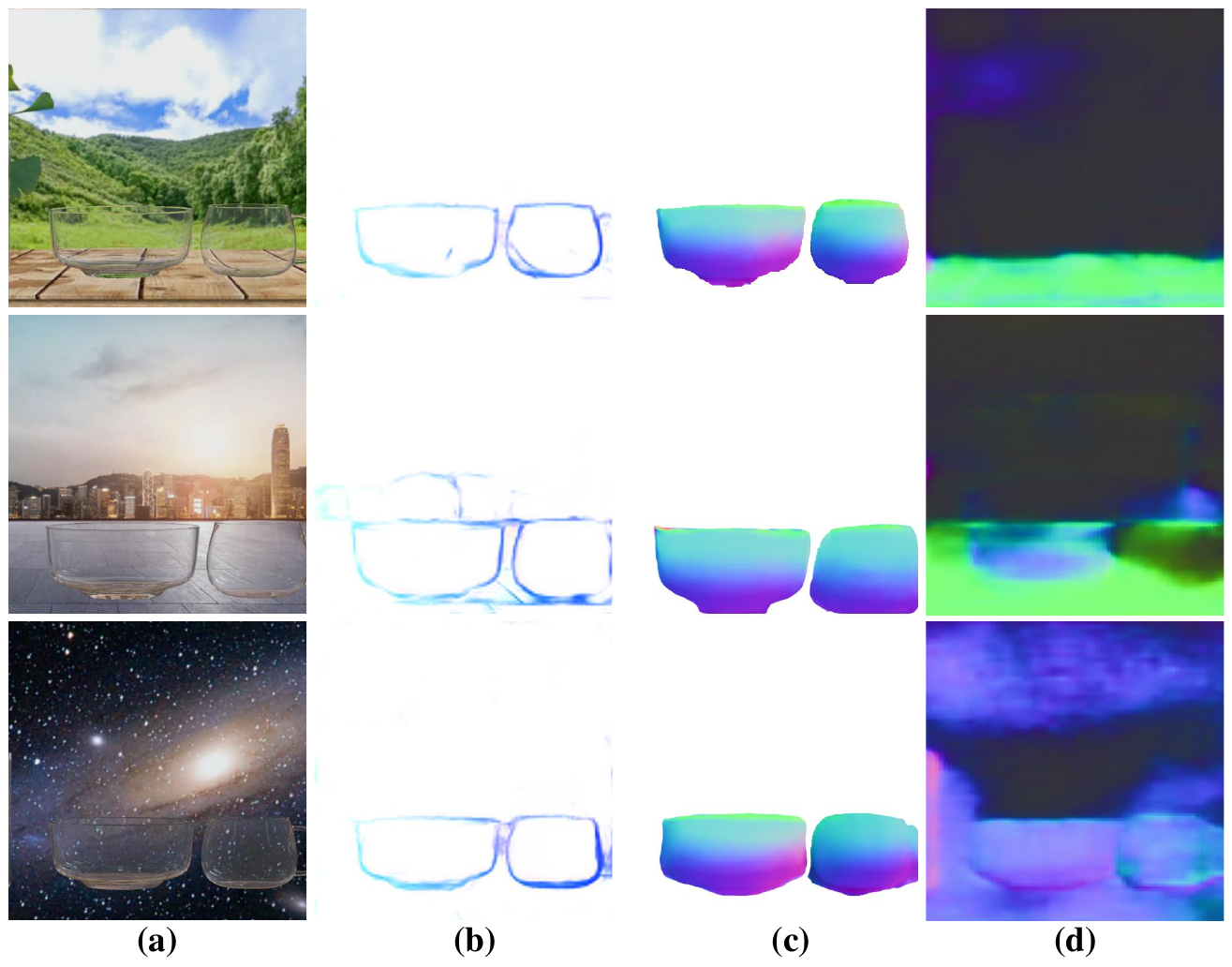}
    \caption{\textbf{(a):} The same transparent objects are placed in front of different, complex backgrounds. \textbf{(b):} The refractive flow predicted by RFNet. \textbf{(c):} The surface normal derived from the refractive flow. \textbf{(d):} The direct surface normal prediction from RGB images \ie ClearGrasp~\cite{cleargrasp}. Please note that RFTrans only predicts the surface normal of transparent objects, while ClearGrasp also predicts the background's normal.}
    \label{fig:rf_matting}
\end{figure}

Refractive flow models the pixel-to-pixel correspondence between the foreground and background, making it possible to handle the cases even when the background texture is complex. As shown in Figure~\ref{fig:rf_matting}, the surface normal from a refractive flow can still be rather stable with a complex background, while the surface normal predicted directly from RGB images as in~\cite{cleargrasp} is not.

\subsection{Geometry Estimation for Transparent Objects}
\label{sec:method_pipeline}

As shown in Figure~\ref{fig:pipeline}, taking in a single RGB-D image, \textbf{RFNet} first predicts the mask, boundary, and refractive flow for transparent objects. Next, the refractive flow is fed into another smaller, \textbf{F2Net} (\textbf{F}low-to-\textbf{N}ormal-Network) which estimates the surface normal. Following previous works~\cite{cleargrasp,a4t}, RFNet is composed by the DeepLabv3+ networks~\cite{deeplabv3} with the DRN-D-54 backbone~\cite{drnbackbone}. The output head is connected to the final feature layer, so the output size is identical to the input size. To note, although the RFNet can consist of any kind of CNN with the encoder-decoder structure, we stick with DeepLabv3+ because we want to keep the network architecture the same as the baseline method for a fair comparison.
We use cross-entropy loss $\mathcal{L}_{CE}$ for mask and boundary and mean squared-error loss $\mathcal{L}_{flow}$ for refractive flow in the training process.
Then, in the F2Net, since refractive flow already encodes rich information about the geometry, we can use a much smaller network like the original U-Net~\cite{RFB15a} to estimate the surface normal from the predicted refractive flow. We use the cosine-similarity loss $\mathcal{L}_{norm}$ to supervise the surface normal. The loss is calculated over the pixels representing the transparent objects. Apparently, the two networks can be jointly optimized, which we will discuss in Section~\ref{sec:ablation_optimization}.

\subsection{Synthetic Dataset Construction}
\label{sec:method_dataset}

\begin{figure}
    \centering
    \includegraphics[width=\columnwidth]{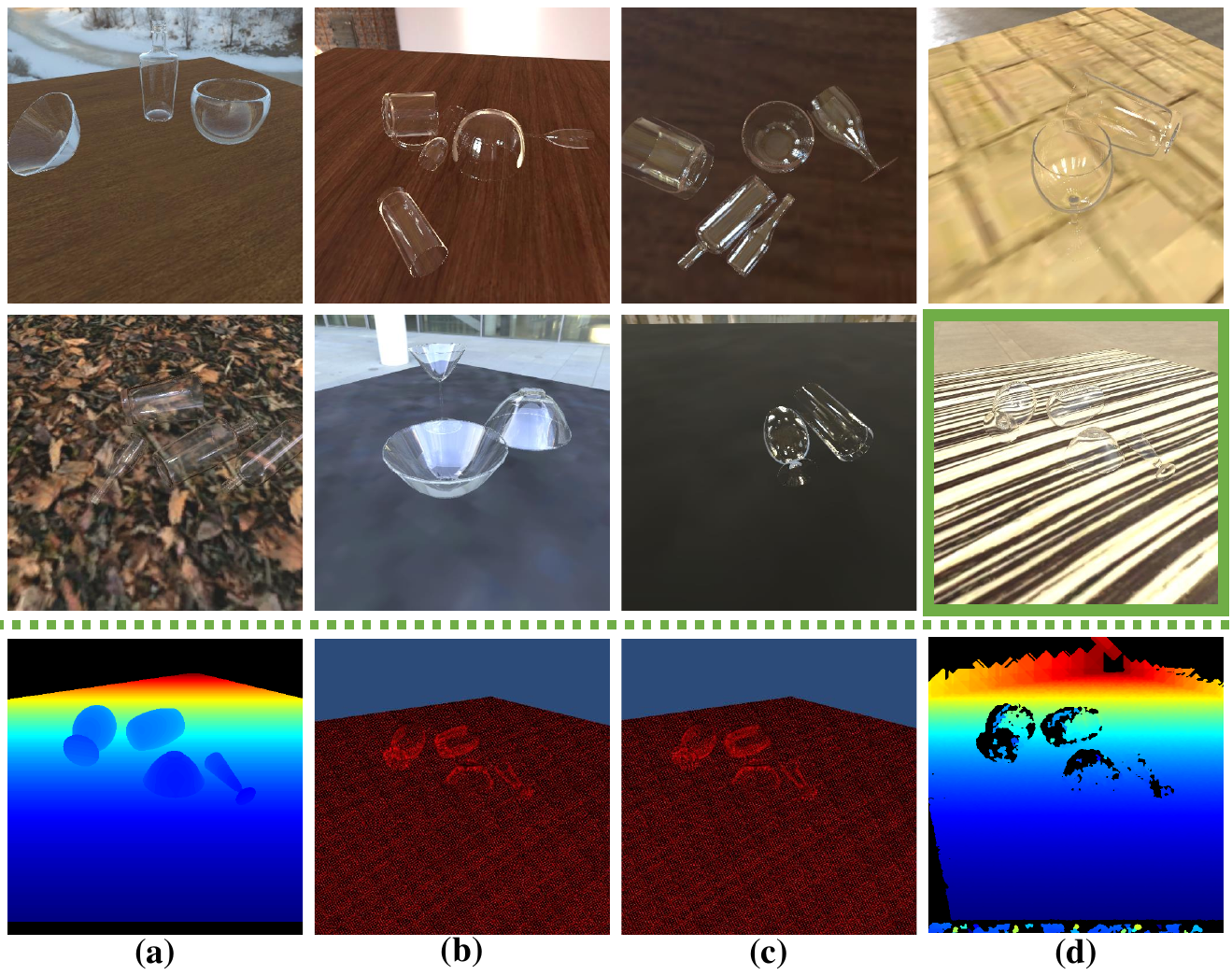}
    \caption{\textbf{Top:} Samples of our synthetic dataset. \textbf{Bottom:} For the RGB image in the green frame (last image in the second row), \textbf{(a)} shows its ideal depth from the depth buffer, \textbf{(b)} and \textbf{(c)} show the simulated left and right IR images, and \textbf{(d)} shows the active depth generated by RFUniverse.}
    \label{fig:example_syn_data}
\end{figure}

We use RFUniverse~\cite{rfu} to construct the synthetic training data.
RFUniverse features accurate, physics-based rendering, which is confirmed in Figure~\ref{fig:sim-to-real-refractive-flow}.
Figure~\ref{fig:example_syn_data} shows some randomly chosen samples of our dataset. Specifically, we collected 438 HDR sky-boxes and 40 textures for the tables. As for object models, the quality of the current large-scale object dataset~\cite{shapenet2015} varies significantly from object to object, where many objects lack internal structures. Other datasets~\cite{fang2022transcg,akb48} built by scanning real objects usually face the problem of wrecks and uneven surfaces. Such defects can lead to unrealistic reflective and refractive effects, making them unsuitable for synthetic dataset construction. By carefully examining the surface quality, geometry, and the resulting reflective and refractive effects, we manually select a total of $62$ different CAD models of $5$ categories, including glass bottles, wine glasses, glass cups, bowls, and plates. To generate one data sample, we first randomly select the sky-box and the texture of the table. Then, we randomly pick 1 to 5 objects and drop them onto the table. The object poses, surface normal, segmentation masks, and boundaries are recorded. As for depth, we collect the active infrared stereo-based synthetic depth~\cite{zhang2023close}, which simulates the Realsense camera and serves as the input. The ideal depth from the depth buffer serves as the ground truth. Finally, we use the gray-code calibration process to get the refractive flow. With one NVIDIA RTX 4090 GPU, the data generation pipeline runs at $\sim$ 5,000 samples per hour, which is quite efficient. The assets are publicly available on our website.

\subsection{Manipulation}
\label{sec:method_grasp}

After the geometry estimation step, we must construct the point cloud of transparent objects for downstream manipulation algorithms. The original depth map, along with the mask, boundary, and surface normal, are used as the input to a global optimization as in~\cite{cleargrasp}. The global optimization will finally output the refined, singulated point cloud with normal.
Since grasping is the foundation of prehensile manipulation, we follow the previous works~\cite{cleargrasp,a4t} to build a real-world grasping environment to demonstrate manipulation.
We use ISF~\cite{isf} as a standalone and efficient grasp planner that can work with grippers of different DoFs. Taking in the singulated point cloud, the ISF will generate valid grasp poses and the corresponding energy-based metric. A lower energy suggests a better grasp pose, so we pick the one with the lowest energy.
In our experiment, we adopt the top-down manner with a Franka Emika panda robot and its original gripper.

%%%%%%%%%%%%%%%%%%%%%%%%%%%%%%%%%%%%%%%%%%%%%%%%%%%%%%%%%%%%%%%%%%%%%%%%%%%%%%%%

\section{Experiment}

To evaluate RFTrans, we test it on both our proposed synthetic dataset, and the real-world benchmark proposed in ClearGrasp~\cite{cleargrasp}. ClearGrasp is considered the baseline model and ImplicitDepth~\cite{implicit_depth} is the previous \textit{state-of-the-art} depth completion method. To prove that the RFTrans trained on synthetic datasets can be directly transferred into real-world applications, we design real-world transparent objects grasping environment and report success rates.

\subsection{Surface Normal Estimation}
\label{sec:experiment_tab_1}
Following previous works~\cite{a4t,cleargrasp}, we evaluate the surface normal estimation performance by three kinds of metrics:  the mean and median errors in degrees, and the percentages of pixels with errors below certain thresholds $\theta$ of 11.25, 22.5, and 30 degrees. Unless specified, metrics are calculated only over the pixels representing transparent objects.

The networks are trained with batch size 16 on a single RTX Titan GPU. The input and output sizes are fixed to 256 by 256. We use SGD optimizer with momentum 0.9 and weight decay 5e-4. The learning rate is fixed to 1e-4 during the whole 100 epochs. It takes less than 48 GPU hours to train the networks.

In Table~\ref{tab:comparison_normal}, test set \textit{Syn} indicates the models are trained on our synthetic dataset of 5,000 images and tested on the synthetic test set of 1,000 images. The proposed method achieves the best result on all metrics. ImplicitDepth is an end-to-end depth completion method, which does not predict surface normal directly, but calculates the normal from the predicted depth (i.e., point cloud). This may explain why it does not perform well. We also measure the average speed. On a single RTX Titan GPU, our method runs at 31 fps. In comparison, ClearGrasp runs at 33 fps and ImplicitDepth runs at 10 fps.

To evaluate the sim-to-real performance, we benchmark the proposed method on the \textit{real-known} test set proposed in ClearGrasp. Instead of using the synthetic train set from ClearGrasp, we have to build our own synthetic train set due to the necessity of refractive flow. Although ClearGrasp does not make its data-generation pipeline open-sourced, it makes the 9 CAD models publicly available, which enables us to rebuild a train set containing 5,000 images of the 9 models. The data generation pipeline is as described in Section~\ref{sec:method_dataset}. In Table~\ref{tab:comparison_normal}, test set \textit{Real} indicates the models are retrained on the synthetic dataset of the 9 objects and are tested on ClearGrasp \textit{real-known} test set. Again, the proposed method achieves the best result on all metrics, which further proves that the refractive flow is a good intermediate representation that can enable a direct sim-to-real transfer.

\begin{table}
\setlength{\tabcolsep}{4pt}
\centering
\caption{Comparison with other methods on surface normal estimation. Test set Syn means the synthetic test set and Real means the real-known test set from the ClearGrasp benchmark. $\downarrow$ suggests the lower the better and vice versa.}
\label{tab:comparison_normal}
\begin{tabular}{ccccccc} \toprule
Test Set& Method& mean$\downarrow$    & med.$\downarrow$    & $\theta_{11.25}\uparrow$ & $\theta_{22.5}\uparrow$   & $\theta_{30}\uparrow$    
\\ \midrule
\multirow{3}{*}{Syn}& \textbf{Ours} & \textbf{11.10}& \textbf{6.94}& \textbf{71.18}&\textbf{85.51}& \textbf{89.18}\\
& ClearGrasp&22.89&17.12&37.56&63.17&72.15    
\\
 & ImplicitDepth& 30.56& 22.10& 23.89& 52.23&63.23\\ \midrule
\multirow{3}{*}{Real}& \textbf{Ours}&\textbf{29.16}&\textbf{24.87}&\textbf{22.94}&\textbf{48.68}&\textbf{61.32} 
\\
& ClearGrasp& 34.76& 30.64& 17.74& 41.54& 53.34 
\\
 & ImplicitDepth& 33.33& 28.75& 12.17& 36.61&53.15\\
 \bottomrule
\end{tabular}
\end{table}

\subsection{Depth Completion}
\label{sec:experiment_depth}
We benchmark the depth completion result to see if the improved surface normal estimation can result in better depth completion in the real world. \ralhigh{For fair comparison, all models are trained on the proposed synthetic dataset} and tested on ClearGrasp \textit{real-known} test set. Following previous works~\cite{cleargrasp,implicit_depth}, we evaluate the depth completion performance by four kinds of metrics: the root mean squared error (RMSE), the absolute relative difference (REL), the mean absolute error (MAE) and the percentages of pixels with errors below certain thresholds $\delta$ of $1.05$, $1.10$, and $1.25$. In Table~\ref{tab:comparison_depth}, results show that \ralhigh{the proposed method can produce the lowest mean error and it performs significantly better than others on the most strict $\delta_{1.05}$ metric.}

\ralhigh{To note, the authors of ClearGrasp report its performance with a heavy pretraining process on 80k images from the Scannet~\cite{dai2017scannet} and Matterport3D~\cite{Matterport3D} datasets in the paper~\cite{cleargrasp}. The proposed method can produce comparable results with only 5k images for training, which proves that introducing refractive flow as the intermediate representation can help make the model significantly less data-hungry. Also, the original paper of ImplicitDepth reports that it can outperform ClearGrasp with its Omniverse Object training dataset of 60k images, while in our experiment with only 5k images, it cannot. This may indicate the size of training set is important for ImplicitDepth.}

\begin{table}
\setlength{\tabcolsep}{4pt} % adjust the space between columns in tables
\centering
\caption{Comparison with other methods on depth completion. $\downarrow$ suggests the lower the better and vice versa. All models are retrained \ralhigh{on the synthetic dataset} and tested on the ClearGrasp real-known benchmark for fair comparison.}
\label{tab:comparison_depth}
\begin{tabular}{ccccccc} \toprule
 Method&  RMSE$\downarrow$&REL$\downarrow$& MAE$\downarrow$& $\delta_{1.05}\uparrow$& $\delta_{1.10}\uparrow$& $\delta_{1.25}\uparrow$\\ \midrule
 Ours&  \textbf{0.038}&\textbf{0.059}& \textbf{0.031}& \textbf{71.46}&\textbf{89.69}& 96.93\\
 ClearGrasp&  0.045&0.071& 0.036& 53.23& 79.01&95.94\\
 ImplicitDepth&  0.048&0.080& 0.043& 28.79& 62.55& \textbf{98.45}\\\bottomrule
\end{tabular}
\end{table}

\subsection{Ablation Study}

In this section, we evaluate the effect of adopting refractive flow, analyze the effects of jointly optimizing the RFNet and F2Net, and check the sensitivity of different viewpoints.

\subsubsection{Refractive Flow}

To evaluate the effect of refractive flow as the intermediate representation, we design three different networks that predict surface normal from different input sources. In Table~\ref{tab:ablation_refractive_flow}, source \textit{RGB} means we directly use the RFNet to predict the surface normal from RGB input, just like what ClearGrasp does. It achieves the worst result. Source \textit{Boundary} means we use the boundary of the objects as the intermediate representation. To be specific, we use RFNet to first predict the boundary, followed by the modified F2Net to predict surface normal. It can be treated as the variant of RFTrans. Source \textit{Flow} means the proposed RFTrans. Results show the boundary can also act as the intermediate representation for surface normal prediction, for the related model outperforms ClearGrasp slightly. Refractive flow achieves the best, which proves that it can encode more information than the plain boundary.

\begin{table}
\centering
\caption{Ablation study on different input sources to estimate surface normal. $\downarrow$ suggests the lower the better and vice versa.}
\label{tab:ablation_refractive_flow}
\begin{tabular}{cccccc}
\toprule
Source& mean$\downarrow$ &med.$\downarrow$    & $\theta_{11.25}\uparrow$ & $\theta_{22.5}\uparrow$   &$\theta_{30}\uparrow$    
\\ \midrule
RGB& 22.89&17.12& 37.56& 63.17&72.15    
\\
Boundary& 21.98& 16.12& 41.24& 65.53&73.94\\
Flow& \textbf{11.10}&\textbf{6.94}& \textbf{71.18}& \textbf{85.51}&\textbf{89.18}\\ \bottomrule
\end{tabular}
\end{table}

\subsubsection{Joint Optimization}
\label{sec:ablation_optimization}
In surface normal estimation via refractive flow, the RFNet and the F2Net can be optimized jointly or separately. Recall that separately, we use $\mathcal{L}_{flow}$ for refractive flow and $\mathcal{L}_{norm}$ for surface normal. In the joint manner, we use $\mathcal{L} = \alpha \mathcal{L}_{flow} + \mathcal{L}_{norm}$ as the final loss. $\alpha$ is the coefficient to balance the two terms.
Here, we quantitatively evaluate the effectiveness of joint optimization of RFNet and F2Net. The networks are trained and tested on the synthetic dataset. Table~\ref{tab:ablation_joint_optimization} shows the joint optimization (denoted by End2End) can significantly boost the performance, compared with the separate optimization. Further, we also analyze the effects of different values of $\alpha$. We vary the value of $\alpha$ from $0.001$ to $1.5$ in the log scale. Results show that we can slightly benefit from tuning this parameter and $\alpha=0.01$ is the best.

\begin{table}
\centering
\caption{Ablation study on joint optimization and the value of $\alpha$. $\downarrow$ suggests the lower the better and vice versa.}
\label{tab:ablation_joint_optimization}
\begin{tabular}{ccccccc}
\toprule
Method &$\alpha$& mean$\downarrow$ &med.$\downarrow$    & $\theta_{11.25}\uparrow$ & $\theta_{22.5}\uparrow$   &$\theta_{30}\uparrow$    
\\ \midrule
Separate  &-& 18.03&12.29& 56.51& 75.47&81.19\\
End2End&0.001& 12.23&7.79& 68.21& 83.59&87.65\\
End2End&0.01& \textbf{11.10}& \textbf{6.94}& \textbf{71.18}& \textbf{85.51}&\textbf{89.18}\\
End2End&0.1& 11.65& 7.36& 70.08& 84.45&88.24\\
End2End&1.0& 12.33&7.88& 67.86& 83.35&87.53\\
End2End&1.5& 12.5&8.08& 67.75& 83.10&87.29\\ \bottomrule
\end{tabular}
\end{table}

\subsubsection{Sensitivity of Viewpoints}

In order to test the model's sensitivity on different viewpoints, we select 2,000 images from the synthetic train set where the pitch angle of the camera stays between 75 and 90 degrees. The derived subset is named \textit{High-Viewpoint} train set. We also select images with the pitch angle between 30 and 75 degrees to get the \textit{Low-Viewpoint} train set. Similarly, we derive the \textit{Low-Viewpoint} test set. In Table~\ref{tab:ablation_sensitivity_viewpoints}, results show that on the \textit{Low-Viewpoint} test set, the network trained on the \textit{High-Viewpoint} set performs significantly worse than the \textit{Low-Viewpoint} one. This experiment confirms that refractive flow is subjective to viewpoints. \ralhigh{Therefore, in order to get best performance, the training set should cover the viewpoint for inference.}

\begin{table}
\setlength{\tabcolsep}{3pt} % adjust the space between columns in tables
\centering
\caption{Sensitivity of viewpoints. $\downarrow$ suggests the lower the better and vice versa.}
\label{tab:ablation_sensitivity_viewpoints}
\begin{tabular}{ccccccc}
\toprule
Train Set&Test Set& mean$\downarrow$ &med.$\downarrow$    & $\theta_{11.25}\uparrow$ & $\theta_{22.5}\uparrow$   &$\theta_{30}\uparrow$    
\\ \midrule
High-View&Low-View& 22.78&18.23& 36.92& 62.68&72.57\\
Low-View&Low-View& 10.74&6.66& 72.09& 86.07&89.62\\ \bottomrule
\end{tabular}
\end{table}

\subsection{Manipulation}

To test RFTrans in real-world manipulation scenarios, we set up a robotic system consisting of a 7-DoF robot arm with a parallel gripper by Franka Emika and an Intel Realsense D415 RGB-D camera. We use the \textit{easy-handeye} software package to calibrate hand-eye. We use the success rate as the evaluation metric. \ralhigh{We collect 10 glass objects for grasping, one of which is color tinted.} We set up 10 scenes for evaluation. In each scene, we randomly place 3 to 5 transparent objects. In each attempt, the robot should pick up an object according to the grasp pose generated by the grasp planning algorithm, ISF~\cite{isf}. An attempt is regarded as successful if the object is lifted at least 20 cm above the table. We shift to a new scene once all objects in the current scene are picked up, or a maximum of 5 trails is reached. Table~\ref{tab:success} reports the success rate of our method compared with the baseline methods. Method \textit{Raw Realsense} means we use the raw depth value to directly construct the point cloud that ISF estimates grasp poses based on. Method \textit{ClearGrasp} means the point cloud is generated by ClearGrasp~\cite{cleargrasp}. With the help of RFTrans, the success rate is increased from 35\% to 83\%, outperforming the baseline method.
There are numerous factors leading to the failure of grasping. First, most transparent objects are composed of glass - a material renowned for its smoothness that often results in grippers losing hold of the object.
\ralhigh{Second, the global optimization algorithm~\cite{cleargrasp} may fail to accurately estimate the depth of the parts not in contact with the table}, such as the stem of a wine glass, leading to incorrect point clouds and subsequent grasping failures.
Lastly, the gripper may contact with the object prematurely as it approaches, causing it to shift and ultimately leading to failure. This issue is related to the trajectory planning algorithm of the robot. Please refer to the supplementary video for details.

\begin{table}
\centering
\caption{Success rate of manipulation.}
\label{tab:success}
\begin{tabular}{cccc}
\toprule
Method   & $\#$trials & $\#$success & Rate \\ \midrule
Raw Realsense& 48& 17& 35\%\\
ClearGrasp& 48& 35& 72\%\\
RFTrans  & 48& 40& 83\%\\ \bottomrule
\end{tabular}
\end{table}

%%%%%%%%%%%%%%%%%%%%%%%%%%%%%%%%%%%%%%%%%%%%%%%%%%%%%%%%%%%%%%%%%%%%%%%%%%%%%%%%

\section{Conclusions, Limitations, and Discussion}

In this paper, we propose a framework that utilizes refractive flow to estimate the geometry and manipulate the transparent objects. Experiments show that refractive flow is a good intermediate representation that can lead to better surface normal estimation, which benefits manipulation. Besides, the networks trained on synthetic data can be used directly for manipulation tasks in the real world, which confirms refractive flow can help the direct sim-to-real transfer.

However, these merits only happen with those commonly used transparent objects featuring thin-shell structures in our daily lives. In an extreme case, with a triangular prism, it's difficult to even get the precise refractive flow due to the dispersion effect. Moreover, our method may fail when objects heavily overlap with each other.

We hope our method can arouse people's interest in utilizing physical properties in this track. Future works can extend this method to more complex objects and tasks or find other interesting physical properties that benefit manipulation. 
Also, since the ISF algorithm supports grippers with high DoFs, our proposed method can be easily extend to a dexterous grasping pipeline.

%%%%%%%%%%%%%%%%%%%%%%%%%%%%%%%%%%%%%%%%%%%%%%%%%%%%%%%%%%%%%%%%%%%%%%%%%%%%%%%%

\bibliographystyle{IEEEtran}
\bibliography{references}

\end{document}